\documentclass[sigconf,natbib=true]{acmart}

\usepackage{bm}
\usepackage{enumitem}
\usepackage{multirow}
\usepackage{graphicx}
\usepackage{subfigure}
\usepackage{bbding}
\usepackage{hyperref}
\usepackage{colortbl}

\definecolor{c1}{HTML}{344C11}
\definecolor{c2}{HTML}{7e0f12}

\newcommand{\model}{\textsc{Mudok}}
\newcommand{\pretrain}{CoPT}
\newcommand{\prompt}{PPT}
\newcommand{\dataset}{KPI}
\newcommand{\cmark}{{\textcolor{c1}\Checkmark}}
\newcommand{\xmark}{{\textcolor{c2}\XSolidBrush}}

\AtBeginDocument{%
  \providecommand\BibTeX{{%
    \normalfont B\kern-0.5em{\scshape i\kern-0.25em b}\kern-0.8em\TeX}}}

\setcopyright{acmlicensed}
\copyrightyear{2018}
\acmYear{2018}
\acmDOI{XXXXXXX.XXXXXXX}

\acmConference[Conference acronym 'XX]{Make sure to enter the correct
  conference title from your rights confirmation emai}{June 03--05,
  2018}{Woodstock, NY}
\acmISBN{978-1-4503-XXXX-X/18/06}

\begin{document}

\title{Multi-domain Knowledge Graph Collaborative Pre-training and Prompt Tuning for Diverse Downstream Tasks}


\author{Yichi Zhang$^{1,2}$, Binbin Hu$^{3}$, Zhuo Chen$^{1,2}$, Lingbing Guo$^{1,2}$, Ziqi Liu$^3$, Zhiqiang Zhang$^3$, Lei Liang$^3$\\  Huajun Chen$^{1,2,4}$ and Wen Zhang$^{1,2}$}
\affiliation{
    \institution{$^1$Zhejiang University \quad $^2$Zhejiang University-Ant Group Joint Laboratory of Knowledge Graph\\ $^3$Ant Group\quad $^4$Alibaba-Zhejiang University Joint Institute of Frontier Technology\\}
    \city{Hangzhou, Zhejiang}
    \country{China}
}
\email{{zhangyichi2022, zhang.wen}@zju.edu.cn}
\renewcommand{\shortauthors}{Yichi Zhang et al.}

\begin{abstract}
    Knowledge graphs (KGs) provide reliable external knowledge for a wide variety of AI tasks in the form of structured triples. \textbf{Knowledge graph pre-training (KGP)} aims to pre-train neural networks on large-scale KGs and provide unified interfaces to enhance different downstream tasks, which is a key direction for KG management, maintenance, and applications. Existing works often focus on purely research questions in open domains, or they are not open source due to data security and privacy in real scenarios. Meanwhile, existing studies have not explored the training efficiency and transferability of KGP models in depth. To address these problems, We propose a framework {\model} to achieve multi-domain collaborative pre-training and efficient prefix prompt tuning to serve diverse downstream tasks like recommendation and text understanding. Our design is a plug-and-play prompt learning approach that can be flexibly adapted to different downstream task backbones. In response to the lack of open-source benchmarks, we constructed a new multi-domain KGP benchmark called {\dataset} with two large-scale KGs and six different sub-domain tasks to evaluate our method and open-sourced it for subsequent research. We evaluated our approach based on constructed KPI benchmarks using diverse backbone models in heterogeneous downstream tasks. The experimental results show that our framework brings significant performance gains, along with its generality, efficiency, and transferability.
\end{abstract}

\ccsdesc[500]{Information systems~Information systems applications}
\ccsdesc[300]{Information systems~Collaborative filtering}
\ccsdesc[300]{Computing methodologies~Natural language processing}

\keywords{Knowledge Graphs, Pre-training and Prompt Tuning, Recommender Systems, Text Understanding}


\maketitle

\section{Introduction}

Knowledge graphs (KGs) encapsulate world knowledge as structured triples in the forms of \textit{(head entity, relation, tail entity)}, signifying a relation between the two entities. This succinct and expressive storage format makes KG as the infrastructure for a plethora of AI tasks such as question answering \cite{QAGNN, KnowPAT}, recommendation \cite{KGAT}, and multi-modal understanding \cite{MMKG-survey}.

\par The application of KG in specific domains is particularly noteworthy as numerous domain-specific KGs are constructed for many industry scenarios such as E-commerce \cite{K3M}, telecom network \cite{telebert}, healthcare \cite{KG-healthcare} for practical, on-the-ground applications. In these contexts, \textbf{knowledge graph pre-training (KGP)} plays a pivotal role. KGP aims to pre-train neural networks on the large-scale KGs to capture the semantic information within the designed model, thereby providing flexible services for downstream tasks such as recommendation systems (RS) \cite{PKGM}, knowledge graph reasoning (KGR), and other NLP tasks. This "pre-train then apply" paradigm has garnered attention in the KG field, sparking a wealth of related research in both industrial and academic communities.

\begin{table*}[]
\caption{An intuitive comparison among existing KGP works and ours. We summarize the shortcomings of the existing work from several perspectives and propose our solution. The following is an abbreviated list of downstream tasks: knowledge graph reasoning (KGR), recommendation systems (RS), and natural language processing (NLP).}
\label{table::introduction}
\resizebox{\textwidth}{!}{
\begin{tabular}{c|cccccccccc}

\toprule

\multirow{2}{*}{\textbf{Research}} & \multirow{2}{*}{\textbf{KG Type}} & \multirow{2}{*}{\textbf{KG Size}} & \multirow{2}{*}{\begin{tabular}[c]{@{}c@{}}\textbf{Open-source}\\ \textbf{Benchmarks}\end{tabular}} & \multicolumn{3}{c}{\textbf{Downstream Tasks}} & \multirow{2}{*}{\begin{tabular}[c]{@{}c@{}}\textbf{Downstream}\\ \textbf{Paradigm}\end{tabular}} &\multirow{2}{*}{\begin{tabular}[c]{@{}c@{}}\textbf{Unified}\\ \textbf{Prompt}\end{tabular}} & \multirow{2}{*}{\textbf{Efficient}} & \multirow{2}{*}{\textbf{Trasferable}} \\
 &  &  &  & \textbf{KGR} & \textbf{RS} & \textbf{NLP} &  &  &  \\
 \midrule
ScoP \cite{KGP-ScoP} & Open-domain & Tiny & \cmark & \cmark & \xmark & \xmark & Fine-tune & - & \cmark & \xmark \\
PKGM \cite{PKGM} & Specific-domain & Large & \xmark & \xmark & \cmark & \xmark & Stand-up & - & \xmark & \xmark \\
KGTransformer \cite{KGtransformer} & Open-domain & Middle & \cmark & \cmark & \xmark & \cmark & Prompt Tuning & \xmark & \cmark & \cmark \\
Our Work & Specific-domain & Large & \cmark & \xmark & \cmark & \cmark & Prompt Tuning & \cmark & \cmark & \cmark \\
\bottomrule
\end{tabular}
}
\end{table*}

\par However, existing KGP researches \cite{PKGM, KGP-ScoP, KGP-KGC, KGtransformer} faces numerous challenges, which can be summarized as the following perspectives: \textbf{efficiency, transferability, downstream paradigm, and reproducibility}. The first three are primarily associated with method design, while reproducibility is largely a challenge related to benchmark datasets. We have outlined the problems and limitations of various existing research works in Table \ref{table::introduction}. Industry-proposed work \cite{PKGM, PKGM2, PKGM3} often trains structured shallow embeddings for each entity and relation, resulting in a lack of transferability and efficiency. Their downstream application paradigm is typically a vanilla stand-up embedding, which doesn't fully leverage the capabilities of pre-trained models. As they use data from real scenarios, these datasets are often not open-sourced due to commercial reasons, making it difficult for subsequent researchers to reproduce their results. Conversely, while academic work \cite{KGtransformer, KGP-KGC, KGP-ScoP} tends to be more innovative in method design, the open-source datasets they use are generally smaller, and the tasks (e.g., KGR) they address are often more idealized and detached from practical applications, lacking research on KGP enhanced recommendation tasks. 

\par To address both the methodological and benchmark challenges mentioned above, we seeks to strike a balance balance between the two types of research. We present a more practical scenario for KGP applications, \textbf{innovative pre-training and downstream task adaptation methods, and new open-source reproducible datasets}. We first propose a research scenario that is more relevant to \textbf{real-world applications}, specifically, enhancing diverse and heterogeneous downstream tasks with a large-scale KG containing multi-domain knowledge. For such a scenario, we propose a \underline{\textbf{Mu}}lti-\underline{\textbf{Do}}main \underline{\textbf{K}}nowledge graph pre-training framework ({\model} for short). {\model} consists of a collaborative pre-training module ({\pretrain} for short) and a prefix prompt tuning module ({\prompt} for short). {\pretrain} is designed to achieve efficient and transferable pre-training on large-scale multi-domain KGs using a transformer \cite{DBLP:conf/nips/transformer} architecture. {\prompt} aims to provide \textbf{unified interface} to support different downstream tasks with prefix prompts on the frozen pre-trained models. This can be easily integrated into different task backbones and eliminates the need for task-specific prompt design like in existing works \cite{KGtransformer}.
Additionally, we develop a large-scale benchmark called \underline{\textbf{K}}nowledge graph \underline{\textbf{P}}re-training as \underline{\textbf{I}}nfrastructure benchmark ({\dataset} for short) using the open-source Amazon \cite{Amazon} and Douban \cite{douban1} product data with two large-scale KGs and six different domain tasks including recommendation and text understanding. We conduct comprehensive experiments on the {\dataset} benchmark to investigate the design of {\model} across diverse downstream tasks and domains, with additional specialized studies focusing on the efficiency and transferability of our method. Our contribution in this paper can be summarized as:

\begin{itemize}
    \item \textbf{Identification of New Problems.} We perform a comprehensive comparison among existing KGP works and analyze their limitations. We propose a new problem scenario for studying KGP applications on multi-domain and heterogeneous tasks with large-scale KGs.
    \item \textbf{Methodology design}. We introduce a new framework {\model} for collaborative pre-training and prefix prompt tuning. This serves different downstream tasks uniformly while maintaining efficiency and transferability across domains.
    \item \textbf{Benchmark construction.} We establish a new large-scale and open-source {\dataset} benchmark to evaluate our method and contribute to reproducible AI research.
    \item \textbf{Comprehensive experiments.} We conduct extendtive experiments on {\dataset} benchmark to demonstrate the effectiveness of {\model} against existing baselines. We further explore to validate the transferability, reasonability, and efficiency of our method.
\end{itemize}

\section{Related Works}
\subsection{Knowledge Graph Pre-training}
Knowledge graph pre-training (KGP) aims to pre-train neural models on the existing KGs and provide semantic-rich knowledge representations to enhance diverse downstream applications. The downstream tasks can be divided into two main categories, in-KG tasks and out-KG tasks. In-KG tasks include knowledge graph completion \cite{KGP-ScoP} and complex query answering \cite{KGP-KGC}, focusing on the structural reasoning of existing KGs. Out-KG tasks primarily expect to draw on pre-trained knowledge representations to complete various types of knowledge-sensitive downstream tasks like text understanding \cite{KBERT}, question answering \cite{KGtransformer, K3M} and recommendation \cite{PKGM, PKGM2, PKGM3}.

For example, some work \cite{KGP-ScoP, KGP-KGC} pre-train  KGs to capture the structural information and enhance the in-KG reasoning tasks (e.g. knowledge graph completion \cite{KGP-ScoP}, complex query answering \cite{KGP-KGC}). ULTRA \cite{KGFound} attempts to build a foundation model for KG reasoning by pre-training on large-scale KGs. KGTrasnformer \cite{KGtransformer} achieves structural knowledge transfer by pre-training and prompt tuning on open-domain KGs with transformer-based \cite{DBLP:conf/nips/transformer} models, serving both in-KG reasoning task and out-KG application tasks.

\par In the industry, KGP also raises high attention. PKGM \cite{PKGM, PKGM2, PKGM3} constructs item KGs and learns structured shallow item embeddings to serve downstream e-commerce tasks like item classification, alignment, and recommendation. K3M \cite{K3M} pre-trains on multi-modal item KGs and employs transformer-based modality encoders \cite{BERT} to capture the multi-modal item representations and enhance downstream tasks.

\subsection{Knowledge-aware Recommendation}
KGs are widely used to enhance current recommender systems \cite{wang_knowledge_2017_survey1} by modeling the side information of items or users. Existing knowledge-aware recommendation approaches can be categorized into three types: embedding-based, path-based, and GNN-based. 
\par Embedding-based methods \cite{EMB-KGREC1, EMB-KGREC2, EMB-KGREC3, EMB-KGREC4} introduce structural KG embeddings into recommendation models to enhance the user and item representations. Path-based methods \cite{PATH-KGREC1, PATH-KGREC2, PATH-KGREC3} explore to discover the semantic information in the relational meta-paths of the side KGs to enhance the recommendation models. GNN-based approaches \cite{KGCN, KGAT, KGIN, KGCL, KGRec, DiffKG} employ GNNs to encode both the user-item interactions and the side KGs and learn knowledge-aware representations for recommendation by knowledge-constrained training objectives, which unifies the previous two paradigms and integrate their strengths.

\par Current research in KG-aware recommendations is oriented towards a single recommendation domain with individual KG. We propose a new paradigm of collaborative pre-training on large-scale KGs across multiple domains, followed by prompt tuning under different scenarios to serve diverse downstream tasks.

\subsection{Knowledge-aware Text Understanding}
KGs are also employed in many knowledge-sensitive text understanding and inference tasks \cite{KGNLP-survey, KGNLP-Survey2}, providing stable and reliable external knowledge. Some works like K-BERT \cite{KBERT}, JAKET \cite{JAKET}, and DRAGON \cite{DRAGON} pre-trains a BERT-like model on large-scale knowledge-aware corpus to serve a series of textual understanding tasks like relation classification \cite{JAKET}, question answering \cite{DRAGON}, entity recognition \cite{KBERT}, and sentiment classification \cite{JAKET}. Some works like QA-GNN \cite{DBLP:conf/naacl/YasunagaRBLL21-QAGNN} and GreaseLM \cite{GreaseLM} combine KGs and language models with knowledge retrieval to answer questions with background knowledge. With the rise of large language models (LLMs) \cite{LLMKG-survey}, more and more studies are considering combining KG and LLM together to achieve knowledgeable LLM application \cite{GNP, KnowPAT}.

\par Our research focuses on a problem of practical value, i.e., how to synergistically pre-train KGs from multiple sub-domains in a large domain (e.g., e-commerce) scenario to serve diverse downstream tasks. We focus on how to combine pre-trained KG models for domain-sensitive NLP tasks such as domain-specific question answering and review rating prediction.

\section{Problem Definition}
\begin{figure*}[]
  \centering
\includegraphics[width=\linewidth]{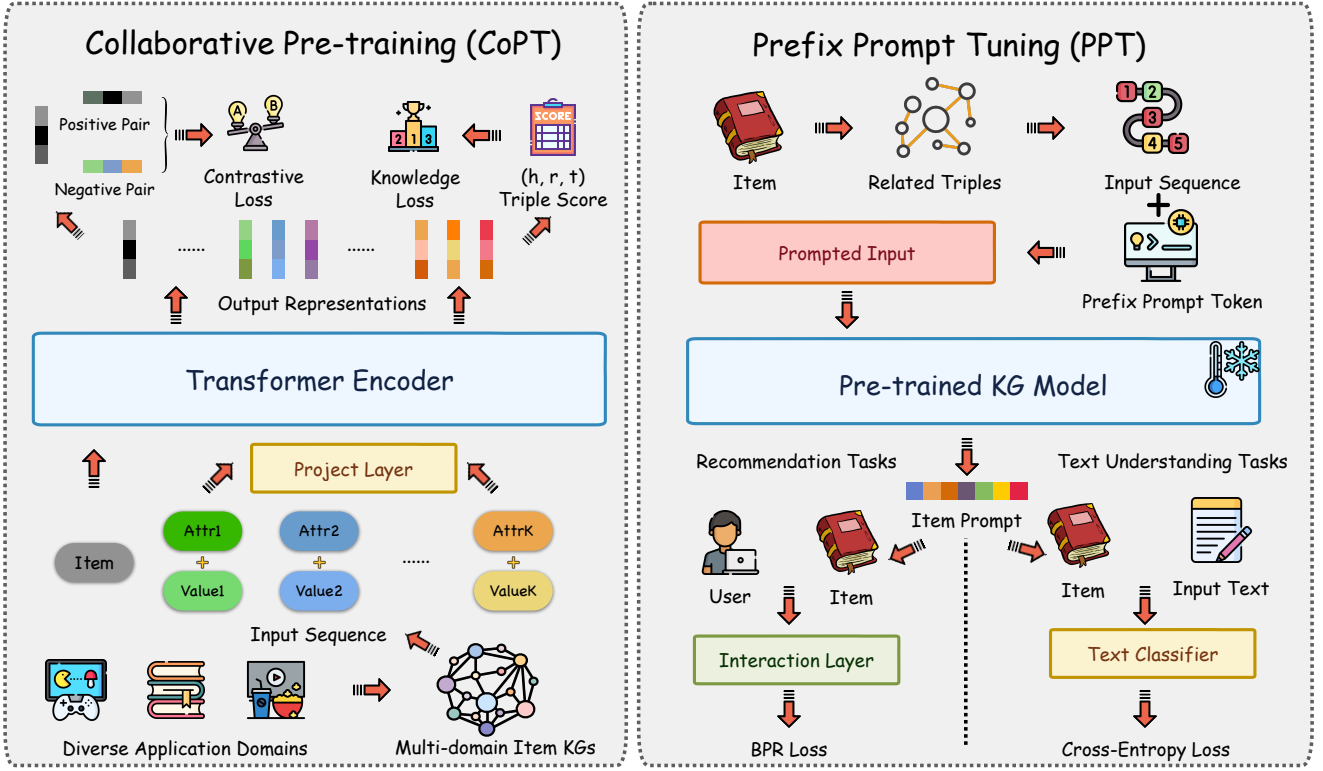}
  \vspace{-8pt}
  \caption{Overview of our proposed framework {\model}. {\model} consists of a collaborative pre-training stage and a prefix prompt tuning stage, which first pre-trains on the large-scale multi-domain item KGs and fine-tuned on the item-aware downstream tasks like recommendation and text understandings with a lightweight prefix prompt token.}
  \label{figure::model}
  \vspace{-8pt}
\end{figure*}

In this section, we introduce the basic notations and problem setting for our scenario: multi-domain KG serves diverse downstream tasks.
\subsection{Multi-domain KGs}
In this section, we begin with a brief introduction to the problem scenarios we propose and the basic notation we used. A general KG can be represented as $\mathcal{KG}=(\mathcal{E}, \mathcal{R}, \mathcal{T})$ where $\mathcal{E}, \mathcal{R}$ are the entity set and the relation set. $\mathcal{T}=\{(h, r, t)\mid h, t \in\mathcal{E}, r\in\mathcal{R}\}$ is the triple set. Each triple $(h, r, t)$ represents a relation $r$ between the head entity $h$ and the tail entity $t$. In this paper, we consider a large-scale item KG as the union of several sub-KGs of multiple different domains of items, which is constructed with massive meta-data of the items in the form of $(item, attribute, value)$. For example, an electronic entertainment item KG can be made up of three different domains: music, film and TV, and games. Each domain has its own KG deposited between them as a data asset, which will be used in different downstream domain-specific tasks. 

\par More formally, we denote the number of multiple domains as $k$. Therefore, the entity set can be split into $k$ domain entity sets as $\mathcal{E} = \mathcal{E}_1 \cup \mathcal{E}_{2} \cup \cdots \cup \mathcal{E}_k$ where $\mathcal{E}_{i} \cap \mathcal{E}_{j}$ can be either empty or non-empty set as some of the items or their profiles might appear in multiple domains. Similarly, the relation set can be split as $\mathcal{R}=\mathcal{R}_1 \cup \mathcal{R}_{2} \cup \cdots \cup \mathcal{R}_k$. Since different attribute types and attribute values may appear in more than one domain, after several sub-KGs are \textbf{combined into a large-scale item KG}, there will be some overlapping entities and relations between different sub-KGs. We will consider the different domains together as a larger KG.

\subsection{Tasks Formulation}
\par Meanwhile, these KGs serve several tasks for each domain like recommendation and text understanding tasks. 

\par A classic recommendation task has a user set $\mathcal{U}$ and an item set $\mathcal{V}$. Besides, we have an interaction graph $\mathcal{G}=\{(u, v)\mid u\in\mathcal{U}, v\in\mathcal{V}\}$ denotes that user $u$ and item $v$ has an interaction (e.g. click). With an extra item KG as infrastructure, it is important to note that the item set is the subset of the entity set, i.e., $\mathcal{V}\subset\mathcal{E}$. Therefore, we can obtain background information from the item KG by modeling the complex relational triples in the given KG, which could benefit the recommendation tasks. The KG-enhance recommendation task aims to learn a model $\mathcal{F}(u,v|\mathcal{G}, \mathcal{KG}, \Theta)$ where $\mathcal{F}$ is a model with learnable parameters $\Theta$. The output of the model is a score to discriminate the likelihood that user $u$ has an interaction with item $v$, which can be learned from the existing interaction graph $\mathcal{G}$ and the item KG $\mathcal{KG}$.

\par Besides, a downstream text understanding task aims to classify the text related to the items., e.g. answering a yes/no question about the item or predicting the score of the item reviews. The text understanding dataset can be denoted as $\mathcal{D}=\{(X_i, v_i, y_i)\mid (i=1,2,\cdots, N)\}$ where $X_i, y_i$ are the input text and its label. $v_i\in\mathcal{V}$ is an item. The general purpose is to learn a discriminative model $\mathcal{H}(y|X, v, \mathcal{KG})$ to predict the label (answer, or review score) based on both input text and extra item knowledge.

\par Both tasks outlined above are \textbf{item-centric} tasks, relying on external knowledge from external item KGs to enhance the performance of the model. However, these tasks are heterogeneous. The recommendation task focuses on user-item interaction graphs, whereas text understanding emphasizes unstructured text modeling. These tasks apply external knowledge differently, with recommendation prioritizing structured information and text comprehension focusing on textual information in the KG. Moreover, these sub-KGs across various domains possess shareable knowledge. Therefore, it is vital to model heterogeneous tasks in different domains collaboratively so that sharing of knowledge can be accomplished between each other and providing uniform interfaces for downstream tasks to enhance these tasks is an important topic. To tackle this issue, we suggest a novel collaborative pre-training and prompt tuning framework for multi-domain KGs to cater to a wide range of downstream tasks.

\section{Methodlogy}

In this section, we propose a unified framework to pre-train on the \textbf{Mu}lti-\textbf{do}main \textbf{K}G ({\model}) and provide downstream service for diverse tasks. Our framework consists of two parts: the \textbf{Co}llaborative \textbf{P}re-\textbf{T}raining module ({\pretrain} for short) and the \textbf{P}refix \textbf{P}rompt \textbf{T}uning module ({\prompt} for short), aiming to achieve \textbf{representative, efficient, and transferable} item knowledge pre-training and application. Figure \ref{figure::model} gives an intuitive overview of our work.

\subsection{Collaborative Pre-training}

\subsubsection{\textup{\textbf{Model Architecture}}}
The collaborative pre-training ({\pretrain}) module aims to capture high-quality contextualized entity and relation representations in the given large-scale $\mathcal{KG}$. We capture an initial feature for each entity $e\in\mathcal{E}$ by encoding their textual descriptions with BERT \cite{BERT}, which are denoted as $\bm{e}\in\mathbb{R}^{d}$ respectively. Besides, each relation $r\in\mathcal{R}$ is represented by an embedding $\bm{r}$. Since all our downstream tasks are item-centric, the design of our pre-training module also unfolds in an item-centric manner.
For an item $h$, we collect its related attributes and values from $\mathcal{KG}$ as $(h, r_1, t_1), (h_, r_2, t_2), \cdots, (h, r_n, t_n)$ where $n$ is the triple amount. Then the input sequence for pre-training can be denoted as:

\begin{equation}
    x=(\mathtt{h}, \mathtt{t}_1, \mathtt{t}_2, \cdots, \mathtt{t}_k)\in\mathbb{R}^{(k+1)\times d}
\end{equation}
where $\mathtt h=\bm{h}W_{proj}$ and $\mathtt t_i=\bm{t}_i W_{proj}+\bm{r}_i$. $W_{proj}\mathbb{R}^(d\times d)$ is a project layer to tranform the initial features. Since an input sequence contains one item (head entity) and multiple attribute values (tail entities), we add the relation representation on the tail entities, allowing relation-aware feature extraction.  In the typical transformer-based pre-training model like BERT \cite{BERT}, the input sequence also adds learnable position embedding. However, the main subjects of our research are triples, and there is no clear sequential relationship between different triples like text, they are all equivalent, so our design does not consider adding position embedding in the input sequence.

We then pass the input sequence $x$ through the transformer encoder \cite{DBLP:conf/nips/transformer} for feature encoding. A classic transformer encoder layer consists of a multi-head attention (MHA) module and a feed-forward network (FFN) module. MHA can be denoted as:
\begin{equation}
    \mathtt{MHA}(x)=\mathtt{Concat}(\mathtt{head_1}, \mathtt{head_2}, \cdots, \mathtt{head}_{h})W_o
\end{equation}
where each attention head $\mathtt{head}_i$ is calculated by:
\begin{equation}
    \begin{aligned}
        \mathtt{head}_i=&\mathtt{Attn}(xW_q^{(i)}, xW_k^{(i)}, xW_v^{(i)}) \\
        =&\mathtt{softmax} \left((xW_q^{(i)})(xW_k^{(i)})^T/\sqrt{d_k}\right)xW_v^{(i)}
    \end{aligned}
\end{equation}
In this equation, $W_q, W_k, W_v, W_o\in\mathbb{R}^{d_{m}\times d_{m}}$ are the projection layers to project the input sequence into the $d$-dimension representation space. An FFN module consists of a two-layer MLP with ReLU \cite{DBLP:journals/jmlr/GlorotBB11_relu} as the activation function. Residual connection and layer normalization are also employed in the standard transformer layer to keep stable training. With $L$ layers of such transformer encoder, the final output can be denoted as:
\begin{equation}
    x'=(\widehat h, \widehat t_1, \widehat t_2, \cdots, \widehat t_k)\in\mathbb{R}^{(k+1)\times d}
\end{equation}
These hidden representations of items and values will be further considered in the pre-train losses.

\subsubsection{\textup{\textbf{Pre-training Objectives}}}
Similar to other large-scale data pre-training such as text \cite{BERT}, images \cite{ViT}, and graphs \cite{GraphPrompt}, we utilize self-supervised learning objectives to extract features from extensive KGs. Our aim is to learn to extract the representational features of an item, which are influenced by its attributes and values. With the transformer encoder, the output item representation $\widehat h$ has been attribute-aware in the context the attribute-value sequence. Motivated by classic sequential representation learning methods like SimCSE \cite{SimCSE}, we can design in-batch contrastive learning as:

\begin{equation}
    \mathcal{L}_{con}=-\sum_{i=1}^{|\mathcal{B}|}\log\frac{\exp(\mathbf{cos}(\widehat{h}_i, \widehat{h}_i')/\tau)}{\sum_{j=1}^{\mathcal{B}}\exp(\mathbf{cos}(\widehat{h}_i, \widehat{h}_j')/\tau)}
\end{equation}
where $\mathcal{B}$ is a batch of items, $\mathbf{cos}(\cdot,\cdot)$ is the cosine similarity of two embeddings and $\tau$ is the temperature hyper-parameter. We get two representations $\widehat{h}_i, \widehat{h}_i'$ from the transformer encoder for each item $h_i$ in a batch $\mathcal{B}$ through two forward passes. We use them as positive pairs, then sample negative samples from the batch for contrastive learning. This approach is feasible due to the dropout layer in the transformer, which introduces little noise and yields a slightly different but fundamentally similar item representation after each forward pass.

\par $\mathcal{L}_{con}$ applies contrastive learning on representations of items from a graph perspective. As part of a KG, an item $h$ also needs to adhere to the structured knowledge constraints $(h_i, r_{i, j}, t_{i, j})$ where $j=1,2,\cdots, n$ is the triples related to $h_i$. Consequently, we define a score function $\mathcal{S}(h, r, t)=(\widehat h\odot \bm{r})\times \widehat t$ to measure the plausibility of each triple where $\odot$ is an element-wise product and $\times$ is the dot product of two vectors. $\mathcal{S}$ transforms the plausibility of a triple into a scalar measure. A higher score represents a correcter knowledge triple. We then design a knowledge triple loss to constraint the item embeddings, which is denoted as:
\begin{equation}
    \mathcal{L}_{kg}=-\sum_{i=1}^{|\mathcal{B}|}\sum_{j=1}^n \frac{\exp(\mathcal{S}(h_i, r_{i, j}, t_{i, j}))}{\sum_{k=1}^n \exp(\mathcal{S}(h_i, r_{i, j}, t_{i, k}))}
\end{equation}
We also use in-batch negative sampling, disrupting the original triple to construct negative samples. With this loss, we can push the entity representations to satisfy the structural constraints of the knowledge triples, thus introducing semantic-rich knowledge information \cite{Knowprompt} in them. The final training objective will be:
\begin{equation}
    \mathcal{L}=\mathcal{L}_{con}+\lambda \mathcal{L}_{kg}
\end{equation}
We combine the two losses together by setting a weight hyper-parameter $\lambda$ and the final pre-training objective can be 
\begin{table*}[]
\caption{The detailed information of our {\dataset} benchmark. Our benchmark consists of two datasets Amazon and Douban. For each dataset, we list the statistics of the item KGs, as well as the task form, domain, and dataset size of the three downstream scenarios. Abbreviations used in the table: Recommendation (Rec), Review Prediction (RP), and Question Answering (QA).}
\label{table::dataset}
\resizebox{\textwidth}{!}{
\begin{tabular}{c|ccc|ccccc|ccccc|ccccc}
\toprule
\multirow{2}{*}{\textbf{Dataset}} & \multicolumn{3}{c|}{\textbf{KG}} & \multicolumn{5}{c|}{\textbf{Downstream Task1}} & \multicolumn{5}{c|}{\textbf{Downstream Task2}} & \multicolumn{5}{c}{\textbf{Downstream Task3}} \\ \cline{2-19} 
 & $|\mathcal{E}|$ & $|\mathcal{R}|$ & $|\mathcal{T}|$ & Task & Domain & \#User & \#Item & \#Data & Task & Domain & \#User & \#Item & \#Data & Task & Domain & \#User & \#Item & \#Data \\ \hline
Amazon & 543331 & 3 & 1347226 & Rec & Music & 27452 & 20677 & 165759 & RP & Movie\&TV & - & 3666 & 50000 & QA & Games & - & 1093 & 5856 \\
Douban & 283756 & 14 & 3600938 & Rec & Book & 1715 & 8592 & 104167 & Rec & Movie & 2633 & 20963 & 1196434 & RP & Music & - & 8763 & 12000 \\
\bottomrule
\end{tabular}

}
\end{table*}
\subsection{Prefix Prompt Tuning}
By pre-training on the large-scale multi-domain KG, we obtain a transformer encoder model for extracting a knowledge-aware representation of the items. This model will subsequently be applied to various downstream tasks. Historically, early applications of pre-trained models like BERT \cite{BERT} necessitated fine-tuning the model's entire parameters. However, this method tends to be computationally demanding, particularly for recommendation tasks requiring larger training batch sizes.
\par Therefore, we refer to the classical parameter-efficient prompt tuning \cite{PEFT} approach in NLP models, , introducing a lightweight \textbf{prefix prompt tuning} ({\prompt} for short) for applying our pre-trained models to different downstream tasks. Rather than tuning the full model, {\prompt} freezes the pre-trained model and adds a special prefix token $p\in\mathbb{R}^{d_p}$ for each item $h\in\mathcal{E}$. A project layer $W\in\mathbb{R}^{d_p\times d}$ is employed to transform the prefix token into the representation space pre-trained model. This transformation is subsequently added to the item representations. Therefore, the input sequence of the pre-trained model will be:
\begin{equation}
    x_{ppt}=(\mathtt{h}+p W_p, \mathtt{t}_1, \mathtt{t}_2, \cdots, \mathtt{t}_k)\in\mathbb{R}^{(k+1)\times d}
\end{equation}
This input sequence will be sent into the pre-trained model and we can get the output prompted representation of the item as $\widehat h_{ppt}$. As all our downstream tasks are item-centric, and our pre-trained model is crafted to enrich the representation of items in these tasks, we only assign extra tokens for items to facilitate parameter-efficient fine-tuning. In the following sections, we will discuss how this prompted representation can be utilized to bolster downstream tasks such as recommendation and textual understanding.

\par It is crucial to highlight that {\prompt} can provide is capable of providing a unified item representation enhancement across various types of downstream task models. This approach isn't confined to a specific downstream task backbone model, thereby allowing us to employ a unified representation for different backbone models in the following sections.

\subsubsection{\textup{\textbf{PPT for Recommendation Tasks}}}
For downstream recommendation tasks, we add the prompted representation $\widehat h_{ppt}$ to the item embedding $v$. The feature interaction between users and items is then allowed through the modules set up by the recommendation model, a process that can be represented as:
\begin{equation}
    \bm{u}, \bm{v}=\mathtt{Interaction}(u, v + \widehat h_{ppt}, \mathcal{G})
\end{equation}
The interaction module is often based on the existing user-item interaction graph for feature interaction, e.g., in LightGCN \cite{lightgcn}, this interaction module is a GCN model of several layers. The prediction model $y_{uv}=\mathcal{F}(u,v|\mathcal{G}, \mathcal{KG}, \Theta)=u^Tv$. For downstream task training, we employ the widely used Bayesian Personalized Ranking (BPR) loss to optimize the model parameters as follows:
\begin{equation}
    \mathcal{L}_{rec}=\sum_{(u, v)\in\mathcal{G}}\sum_{(u, w)\notin\mathcal{G}} -\log\sigma(y_{uv}-y_{uw})+\mu ||\Theta||^2_2
\end{equation}
where $(u, v)$ is an observed positive interaction and $(u, w)$ is a sampled negative interaction. $\sigma$ is the sigmoid function and $||\Theta||^2_2$ is the regular term to avoid overfitting, which is controlled by the weight $\mu$. Such an approach follows the classic paradigm of item recommendation, while further enhancing the features of the item through the representations extracted from the pre-trained model, which will further play a role in the complex interaction module of the original recommendation model.

\subsubsection{\textup{\textbf{PPT for Text Understanding Tasks}}}
For downstream text understanding tasks, we can also add the prompted representation to enhance the final prediction. As mentioned before, the dataset of item-related text understanding tasks can be formulated as $\mathcal{D}=\{(X, v, y)\}$. For a base model, the text $i$ and the textual description of item $v$ will be the input to a pre-trained language model (PLM) like BERT \cite{BERT} and obtain the contextualized text representations. This can be denoted as:
\begin{equation}
    O=\mathtt{Pooler}(\mathtt{PLM}(X, v))
\end{equation}
where $\mathtt{Pooler}()$ is the pooling layer of the PLM. We can also add the prompted representation $\widehat h_{ppt}$ on the PLM output and obtain the predicted probability distribution of each label with softmax normalization, which can be denoted as:
\begin{equation}
    P_{c}=\frac{\exp\left(z_c\right)}{\sum_{j=1}^{|\mathcal{C}|}\exp\left(z_j\right)}
\end{equation}
where $\mathcal{C}$ is the label set and $z_c$ is the probability logits of class $c$ predicted by the model. It is generated by another MLP layer parameterized by $(W_c, b_c)$ as:
\begin{equation}
    z_c=(O+\widehat h_{ppt})W_c + b_c
\end{equation}
The final training objective of the downstream textual understanding tasks is the cross-entropy loss for classification:
\begin{equation}
\mathcal{L}_{text}=-\frac{1}{|\mathcal{D}|}\sum_{i=1}^{|\mathcal{D}|}\sum_{c=1}^{|\mathcal{C}|} y_{i,c}\log(P_{i,c})
\end{equation}
where $y_{i, c}$ is the class label of $i$-th data
For different text understanding tasks such as item QA and review score prediction, we use the same framework described above for training and learning the downstream tasks.

\subsection{Efficiency and Transferability Analysis}
With the framework described above, we have built a representative framework {\model} for pre-training item knowledge and applying it for lightweight fine-tuning in item-related downstream tasks. Below we present some details in this {\model} framework to illustrate why our framework is efficient and transferable in the two stages: the pre-training stage and the prompt tuning stage.

\par During pre-training of {\model}, we freeze the features $\bm{e}$ of all entities during pre-training derived from the text encoder. OOur primary objective is to train the transformer encoder and the parameters in the projection layer to \textbf{adapt} to the input from the textual representation space and our training goals. This design has two benefits. Firstly, by keeping the features of entities frozen, we significantly decrease the number of parameters that need to be trained during pre-training, thus enhancing efficiency. Secondly, we use the textual feature space as a conduit to achieve the cross-domain knowledge generalization and transfer.  The transformer encoder is trained to assimilate knowledge from various domains, ensuring good generalization ability even when a new domain KG is added to the large-scale KGs after pre-training. Hence, the pre-training phase guarantees the model's efficiency and transferability.

\par In the prompt tuning stage, we freeze the parameters of the entire pre-training model and only fine-tune a minuscule fraction of the additional parameters. This fraction is significantly smaller than the entire pre-training model. Our {\prompt} is designed to be \textbf{task-agnostic}, with different tasks obtaining a representation from the pre-training model and using it for the downstream task in a consistent manner. We can generalize to different tasks with the same pre-trained model. In essence, our framework {\model} exhibits transferability and efficiency in both pre-training and prompt tuning phases. We will further demonstrate this in our experimental section.
\section{Experiments and Evaluation}
\begin{table*}[]
\caption{Main experiment results for recommendation of {\model} and baseline methods. The best results are bold and the second best results are underlined for each backbone. We report the results of three metric and the improvements that {\model} brings.}
\label{table::recommendation}
\resizebox{0.95\textwidth}{!}{
\begin{tabular}{c|l|ccc|ccc|ccc}
\toprule
\multirow{2}{*}{\textbf{Backbone}} & \multicolumn{1}{c|}{\multirow{2}{*}{\textbf{\begin{tabular}[c]{@{}c@{}}Tuning\\ Method\end{tabular}}}} & \multicolumn{3}{c|}{\textbf{Amazon-Music}} & \multicolumn{3}{c|}{\textbf{Douban-Book}} & \multicolumn{3}{c}{\textbf{Douban-Movie}} \\
 & \multicolumn{1}{c|}{} & \textbf{Recall@5} & \textbf{Recall@20} & \textbf{NDCG@5} & \textbf{Recall@5} & \textbf{Recall@20} & \textbf{NDCG@5} & \textbf{Recall@5} & \textbf{Recall@20} & \textbf{NDCG@5} \\ \midrule
\multirow{4}{*}{\textbf{CF} \cite{cf}} & Base & 0.0397 & \underline{0.0947} & 0.0261 & \underline{0.0321} & 0.0714 & \underline{0.0372} & 0.0249 & 0.0713 & \underline{0.1389} \\
 & PKGM & \underline{0.0398} & 0.0926 & \underline{0.0262} & 0.0290 & \underline{0.0730} & 0.0319 & \underline{0.0258} & \underline{0.0714} & 0.1281 \\
 & \model & \textbf{0.0481} & \textbf{0.1053} & \textbf{0.0314} & \textbf{0.0345} & \textbf{0.0756} & \textbf{0.0373} & \textbf{0.0266} & \textbf{0.0730} & \textbf{0.1458} \\
 & Improve & +17.3\% & +11.2\% & +16.6\% & +6.9\% & +3.6\% & +0.3\% & +3.1\% & +2.2\% & +5.0\% \\ \midrule
\multirow{4}{*}{\textbf{NCF} \cite{ncf}} & Base & 0.0412 & \underline{0.0963} & \underline{0.0275} & 0.0270 & 0.0556 & \underline{0.0298} & 0.0206 & \textbf{0.0670} & 0.1105 \\
 & PKGM & \underline{0.0414} & 0.0956 & 0.0270 & \underline{0.0274} & \textbf{0.0633} & 0.0297 & \underline{0.0207} & 0.0655 & \textbf{0.1159} \\
 & \model & \textbf{0.0432} & \textbf{0.0980} & \textbf{0.0289} & \textbf{0.0283} & \underline{0.0583} & \textbf{0.0343} & \textbf{0.0213} & \underline{0.0649} & \underline{0.1123} \\
 & Improve & +4.3\% & +1.8\% & +5.1\% & +3.3\% & - & +15.1\% & +2.9\% & - & - \\ \midrule
\multirow{4}{*}{\textbf{LightGCN} \cite{lightgcn}} & Base & \underline{0.0487} & \underline{0.1132} & \underline{0.0316} & \underline{0.0350} & \underline{0.0827} & \textbf{0.0414} & \underline{0.0255} & 0.0734 & \underline{0.1433} \\
 & PKGM & 0.0374 & 0.0959 & 0.0251 & 0.0333 & 0.0809 & 0.0383 & 0.0265 & \underline{0.0738} & 0.1428 \\
 & \model & \textbf{0.0600} & \textbf{0.1329} & \textbf{0.0399} & \textbf{0.0357} & \textbf{0.0846} & \underline{0.0406} & \textbf{0.0272} & \textbf{0.0779} & \textbf{0.1499} \\
 & Improve & +23.2\% & +17.4\% & +20.8\% & +2.0\% & +2.3\% & - & +6.7\% & +5.6\% & +4.6\% \\ \midrule
\multirow{4}{*}{\textbf{GCCF} \cite{gccf}} & Base & \underline{0.0460} & \underline{0.1171} & \underline{0.0298} & 0.0335 & 0.0768 & \textbf{0.0389} & \underline{0.0264} & 0.0755 & \underline{0.1463} \\
 & PKGM & 0.0449 & 0.1161 & 0.0288 &  \underline{0.0341} & \underline{0.0788} & \underline{0.0378}  & 0.0259 & \textbf{0.0778} & 0.1462 \\
 & \model & \textbf{0.0499} & \textbf{0.1224} & \textbf{0.0323} & \textbf{0.0348} & \textbf{0.0825} & 0.0373 & \textbf{0.0264} & \underline{0.0755} & \textbf{0.1470} \\
 & Improve & +8.5\% & +4.5\% & +8.4\% & +2.1\% & +4.7\% & - & +0.0\% & - & +0.1\% \\ \midrule
\multirow{4}{*}{\textbf{DCCF} \cite{dccf}} & Base & 0.0689 & 0.1329 & 0.0464 & 0.0304 & \underline{0.0629} & 0.0318 & 0.0205 & \underline{0.0691} & 0.0847 \\
 & PKGM & \underline{0.0724} & \underline{0.1404} & \underline{0.0491} & \underline{0.0316} & 0.0590 & \underline{0.0318} & \underline{0.0209} & 0.0681 & \underline{0.0848} \\
 & \model & \textbf{0.0741} & \textbf{0.1460} & \textbf{0.0504} & \textbf{0.0337} & \textbf{0.0761} & \textbf{0.0358} & \textbf{0.0222} & \textbf{0.0692} & \textbf{0.0912} \\
 & Improve & +2.3\% & +4.0\% & +2.6\% & +6.6\% & +17.34\% & +12.6\% & +6.2\% & +0.1\% & +7.5\% \\ \midrule
\multirow{4}{*}{\textbf{SimGCL} \cite{simgcl}} & Base & \underline{0.0658} & 0.1320 & \underline{0.0440} & 0.0368 & 0.0850 & \underline{0.0415} & 0.0255 & 0.0746 & 0.1424 \\
 & PKGM & 0.0648  & \underline{0.1374}  &  0.0436 & \underline{0.0376} & \textbf{0.0859} & 0.0409 & \underline{0.0263}  & \underline{0.0794} & \underline{0.1501} \\
 & \model & \textbf{0.0691} & \textbf{0.1412} & \textbf{0.0469} & \textbf{0.0414} & \underline{0.0856} & \textbf{0.0450} & \textbf{0.0277} & \textbf{0.0798} & \textbf{0.1533} \\
 & Improve & +5.0\% & +2.7\% & +6.6\% & +10.0\% & - & +8.4\% & +5.3\% & +0.5\% & 2.1\%  \\ \bottomrule
\end{tabular}

}
\end{table*}

\begin{table*}[]
\caption{Main experiment results for text understanding tasks including question answering (QA) and review prediction (RP). The best Acc and F1 results are bold and the second best results are underlined for each backbone.}
\label{table::nlp}
\begin{tabular}{c|l|cccl|cc|cc}
\toprule
\multirow{2}{*}{\textbf{Backbone}} & \multicolumn{1}{c|}{\multirow{2}{*}{\textbf{\begin{tabular}[c]{@{}c@{}}Tuning\\ Method\end{tabular}}}} & \multicolumn{4}{c|}{\textbf{Amazon-Game (QA, 2-class)}} & \multicolumn{2}{c|}{\textbf{Amazon-Movie (RP, 5-class)}} & \multicolumn{2}{c}{\textbf{Douban-Music (RP, 5-class)}} \\
 & \multicolumn{1}{c|}{} & \textbf{Acc} & \textbf{P} & \textbf{R} & \multicolumn{1}{c|}{\textbf{F1}} & \textbf{Macro-F1} & \textbf{Micro-F1} & \textbf{Macro-F1} & \textbf{Micro-F1} \\ \midrule
\multirow{4}{*}{\textbf{BERT} \cite{BERT}} & Base & \underline{75.34} & 71.57 & 84.25 & \underline{77.36} & \underline{55.46} & \underline{55.20} & 32.03 & 33.33 \\
 & PKGM & 72.95 & 68.72 & 84.25 & 75.69 & 54.01 & 54.48 & \underline{32.87} & \underline{33.50} \\
 & {\model} & \textbf{79.28} & 80.00 & 78.08 & \textbf{79.03} & \textbf{56.43} & \textbf{56.96} & \textbf{33.18} & \textbf{35.08} \\
 & Improve & +5.2\% & - & - & +2.2\% & +1.7\% & +3.2\% & +0.9\% & +4.7\% \\ \midrule
\multirow{4}{*}{\textbf{RoBERTa} \cite{roberta}} & Base & \underline{76.88} & 80.31 & 71.23 & \underline{75.50} & \underline{57.62} & \underline{57.80} & 30.67 & 33.50 \\
 & PKGM & 72.43 & 80.18 & 59.59 & 68.37 & 55.93 & 56.12 & \underline{33.49} & \underline{34.75} \\
 & {\model} & \textbf{78.25} & 80.00 & 75.34 & \textbf{77.60} & \textbf{59.42} & \textbf{59.46} & \textbf{34.05} & \textbf{35.75} \\
 & Improve & +1.8\% & - & - & +2.8\% & +3.1\% & +2.9\% & +1.7\% & +2.9\% \\ \midrule
\multirow{4}{*}{\textbf{GPT2} \cite{gpt}} & Base & 70.38 & 70.31 & 70.55 & \underline{70.43} &\underline{54.18} & \underline{54.98} & 30.25 & 33.25 \\
 & PKGM & \underline{71.58} & 78.12 & 59.93 & 67.83 & 52.32 & 53.06 & \underline{31.91} & \underline{34.08} \\
 & {\model} & \textbf{78.08} & 87.61 & 65.41 & \textbf{74.90} & \textbf{55.86} & \textbf{55.68} & \textbf{34.65} & \textbf{36.67} \\
 & Improve & +9.1\% & - & - & +6.3\% & +3.1\% & +1.3\% & +8.6\% & +7.6\% \\
 \bottomrule
\end{tabular}
\end{table*}

In this section, we will present the experiments and evaluation results of the {\model} framework. We first introduce the dataset construction process and basic information about our new benchmark, followed by a comprehensive discussion of extensive experiments to highlight the strengths of our framework across a variety of datasets.  The following four research questions (RQ) are the key questions that we explore in the experiments. 

\begin{itemize} 
    \item[\textbf{RQ1.}] Can our framework {\model} enhance the existing baselines and make substantial progress in diverse downstream tasks?
    \item[\textbf{RQ2.}] Is {\model} a transferable framework that can be generalized to item KGs in new domains?
    \item[\textbf{RQ3.}] How much do each module in the {\model} contribute to the final results? Are these modules reasonably designed?
    \item[\textbf{RQ4.}] How does the training efficiency of {\model} compare to existing baselines? Is it efficient enough?
\end{itemize}

\subsection{Dataset Construction}
This study primarily concentrates on multi-domain KG collaborative pre-training to serve diverse downstream tasks, which is a fervent demand in the industry. However, the research community lacks datasets that are fully compatible with this scenario, and much of the industry's research and associated data cannot be made available due to various constraints. Thus, we aim to develop large-scale open source datasets that satisfy our setting based on the existing open source data like Amazon product data \cite{Amazon} and Douban review \cite{douban1, douban2}, aiming to achieve \textbf{K}nowledge \textbf{P}re-training as \textbf{I}nfrastructure (named as {\dataset} benchmark for short).

\par During the dataset construction, we collect multi-domain KGs from each data source (Amazon and Douban) and integrate them into a large-scale KG. The KG information is derived from the meta information of the item organized them into triples in the form of $(item, attribute, value)$. In each dataset, we considered three domains:

\begin{itemize}
    \item Amazon: Digital Music, Movie \& TV, and Video Games
    \item Douban: Book, Music, and Movie
\end{itemize}
For each domain, we define a subtask for it, including three categories: recommendation, question answering (QA), and review prediction (RP). The recommendation is the classic product recommendation task where we build a dataset based on user interaction records. QA and RP are two text understanding tasks aiming to predict "Yes/No" for a given item-related question or predict the user's rating score (1-5) for the item based on the user's review. We obtain the input data (questions/reviews) and corresponding labels from the original data source ensuring categories balance. Detailed information of our {\dataset} is presented in Table \ref{table::dataset} which includes the statistical information of the item KG and downstream tasks.

\subsection{Experiment Settings}
In this section, we mainly talk about the detailed settings of our experiments from three perspectives: the baseline methods, evaluation metrics, and implementation details.
\subsubsection{\textup{\textbf{Baseline Methods}}}
Based on the previous description, our proposed framework is a set of plug-and-play prompt tuning methods that can be flexibly adapted to various recommendation and NLP backbones. Consequently, we select several classic backbone models to evaluate the performance of {\model} by considering the gains our approach can bring to the backbone. For recommendation tasks, we select 6 classic recommendation methods:

\begin{itemize}
    \item \textbf{CF} \cite{cf}: It is the basic collaborative filtering method.
    \item \textbf{NCF} \cite{ncf}: It employs neural networks to model user-item interactions and enhance the CF process.
    \item \textbf{LightGCN} \cite{lightgcn}: It designs a lightweight GCN model for message passing on user-item graphs.
    \item \textbf{GCCF} \cite{gccf}: It simplifies recommendation models by new GNN internal structure design.
    \item \textbf{DCCF} \cite{dccf}: It learns disentangled user/item representations for better CF by contrastive learning.
    \item \textbf{SimGCL} \cite{simgcl}: It enhances CF model performance with an augmentation-free view generation technique.
\end{itemize}
For text understanding tasks, we employ several transformer-based pre-train language models BERT-base \cite{BERT}, RoBERTa-base \cite{roberta}, and GPT2 \cite{gpt} as the text backbone. Above is our selection strategy for the backbone. Meanwhile, we have adopted the following enhancement settings for each backbone:

\begin{itemize}
    \item Base model. It means just training the model from scratch without any KG pre-training and enhancement.
    \item PKGM \cite{PKGM}. It provides a stand-up item embedding with KG pre-training to serve downstream tasks.
    \item {\model}. This is our prefix prompt tuning framework.
\end{itemize}

\subsubsection{\textup{\textbf{Evaluation Metrics}}}
For recommendation tasks, we use rank-based metrics like Recall@K (K=5, 20), and NDCG@5 \cite{lightgcn} following the standard evaluation protocol. For text understanding tasks, we employ classification metrics to evaluate the models. QA is a binary classification task in our dataset. We report accuracy (Acc), precision (P), recall (R), and F1-score (F1) for the QA task. Review prediction is a 5-label classification task and we report the macro-F1 and micro-F1 results.

\subsubsection{\textup{\textbf{Implementation Details}}}
We implement our framework with PyTorch \cite{pytorch} on a Linux server with the Ubuntu 20.04.1 operating system. All the experiments are running on a single NVIDIA A800 GPU. In the pre-train stage, we set the batch size to 1024 and the embedding dimension of the transformer encoder to 128. The initial features of each entity are extracted with BERT with 768 dimensions. We set the triple number of each training sequence to 8. The constrastive temperature $\tau$ and the loss weight $\lambda$ is searched in $\{0.1, 0.5\}$. We pre-train the large-scale KGs for 5 epochs and employ the default Adam optimizer to optimize the model with a learning rate of $5e^{-4}$. For downstream PPT, we set the prefix token dimension to 16 and the projection layer $W_p$ is a $16 \times 128$ matrix. Also, we have different settings for different downstream tasks like recommendation and text understanding.

\par For downstream recommendation tasks, we employ SSLRec \cite{SSLRec}, an open-source recommendation library to conduct the experiments. The embedding dimension of users and items is fixed to 32, the learning rate of Adam optimizer is set to $1e^{-3}$ and the batch size is fixed to 4096. We trained different recommendation models concerning the parameter configurations provided by SSLRec. For downstream text understanding tasks, we implement the models with huggingface transformers \cite{huggingface-transformers} library. The pooler output of each backbone model is used for the final prediction.
The batch size is fixed to 32 and the learning rate of the Adam optimizer is searched in $\{1e^{-5},3e^{-5}, 5e^{-5}\}$. We train each model with 10 epochs. 
\subsection{Main Results (RQ1)}
The main results of recommendation tasks on three domains of the {\dataset} benchmark are outlined in Table \ref{table::recommendation}. We discuss on the performance of different KG enhancement methods on different backbones and the gains {\model} can bring. Macroscopically, we can find that our method achieves some improvement over most of the datasets and backbones. {\model} shows superior performance on the Amazon Music subtask and achieves significant improvements on all ranking metrics across backbones, indicating its generalizability and ability to achieve gains across diverse backbones. Meanwhile, it can be observed that our method performs relatively worse on NCF, while it performs better on LightGCN and those GNN-based methods. We believe that the reason for such a result is that the GNN-based CF method allows the information of the prompt vector provided by the pre-trained model to be passed over the whole user-item graph as well through the message passing of GNN, thereby amplifying the effect of this prefix prompt. Simultaneously, our approach obtains clear improvements in all Recall@5 metrics, which suggests that {\model} can effectively facilitate the recommendation backbone's precise identification of the nature of the goods, and effectively enhance the precise sorting and recall of the goods.

\par Besides, the experiment results of the text understanding tasks are presented in Table \ref{table::nlp}. These experimental results reveal that {\model} offers a substantial improvement over the baseline across all datasets and NLP subtasks. This signifies that our framework generates a comprehensive item representation for downstream text understanding tasks via pre-training. Furthermore, prompt tuning aids the backbones in comprehending the features inherent in the merchandise, thereby enabling the model to accurately predict the nature of the merchandise and user reviews. It is noteworthy mentioning that BERT/Roberta and GPT are two different types of language models, namely, masked language model and causal language model respectively. Our approach {\model} has shown significant advancements in these diverse language modeling commodities, thereby demonstrating our approach's broad applicability.
\begin{table}[]
\caption{The OOD transfer experiments results on the Amazon dataset of KPI benchmark. The transfer setting A+B $\rightarrow$ C represents training on domain A/B and prompt tuning on domain C. (A: Movie, B: Games, C: Music). We compare the OOD results with {\model} pre-trained on full data (Full) and w/o {\model} (None) for several backbones.}
\label{table::transfer}
\resizebox{0.9\columnwidth}{!}{
\begin{tabular}{c|c|c|cc}
\toprule
\multirow{2}{*}{\textbf{\begin{tabular}[c]{@{}c@{}}Domain\\ Transfer\end{tabular}}} & \multirow{2}{*}{\textbf{Backbone}} & \multirow{2}{*}{\textbf{Setting}} & \multicolumn{2}{c}{\textbf{Metrics}} \\
 &  &  & \textbf{Recall@5} & \textbf{NDCG@5} \\
 \midrule
\multirow{6}{*}{\textbf{A+B$\rightarrow$C}} & \multirow{3}{*}{\textbf{LightGCN}} & Full & 0.0600 & 0.0323 \\
 &  & OOD & 0.0549 & 0.0364 \\
 &  & None & 0.0487 & 0.0298 \\ \cmidrule{2-5} 
 & \multirow{3}{*}{\textbf{SimGCL}} & Full & 0.0691 & 0.0469 \\
 &  & OOD & 0.0673 & 0.0457 \\
 &  & None & 0.0658 & 0.0440 \\
 \midrule
\multicolumn{3}{c}{} & \textbf{Acc} & \textbf{F1} \\
\midrule
\multirow{3}{*}{\textbf{A+C$\rightarrow$B}} & \multirow{3}{*}{\textbf{BERT}} & Full & 79.28 & 79.03 \\
 &  & OOD & 77.23 & 76.46 \\
 &  & None & 75.34 & 77.36 \\
 \midrule
\multicolumn{3}{c}{} & \textbf{Ma-F1} & \textbf{Mi-F1} \\
\midrule
\multirow{3}{*}{\textbf{B+C$\rightarrow$A}} & \multirow{3}{*}{\textbf{BERT}} & Full & 56.43 & 56.96 \\
 &  & OOD & 55.46 & 55.74 \\
 &  & None & 55.46 & 55.20 \\
 \bottomrule
\end{tabular}
}
\vspace{-16pt}
\end{table}
\subsection{Tranferability Experiments (RQ2)}
We have previously assert in the paper that our framework {\model} is transferable and generalizable. This is achieved by setting the initial features of the entities extracted from the texts. To validate the transferability of our methodology, we conducted an \textbf{out-of-domain (OOD) transferability} experiment.

\par Specifically, we explored the three sub-domains of Amazon and their corresponding subtasks. We utilized the KGs from the remaining two domains for model pre-training, subsequently performing prompt tuning on the current domains for their respective tasks. The results are shown in Table \ref{table::transfer}. From the experimental results, we can find that our method has a better transferability. The experimental results on all three sub-domains achieve significantly better results than without pre-training, but they are slightly lower than the results on full data. This indicates that our design facilitates OOD knowledge transfer through the unified representation space provided by textual information.
\begin{figure}[]
  \centering
\includegraphics[width=0.95\linewidth]{pictures/ablation.pdf}
  \vspace{-16pt}
  \caption{The ablation study results on the three domains of Amazon. We design five groups of experiments to validate the effectiveness of our design in {\model}. G1: Full Model; G2: w/o {\prompt}; G3: w/o {\pretrain}; G4: w/o {$\mathcal{L}_{con}$}; G5: w/o {$\mathcal{L}_{kg}$}.}
  \label{figure::abaltion}
  \vspace{-16pt}
\end{figure}
\subsection{Abaltion Study (RQ3)}
To further demonstrate the effectiveness of our design in {\model}, we conduct ablation experiments on the three domains of the Amazon dataset. We verify the necessity of the pre-training module {\pretrain} as well as the design of the losses and the {\prompt} module. The experiment results are demonstrated in Figure \ref{figure::abaltion}.

\par From the experimental results, we can find that both our {\pretrain} and {\prompt} modules contribute to the final performance across three different domains and tasks. There is a significant decrease in the performance of the model when it is directly tuned on the model without pre-training (G3). This is because models that are not pre-trained are very prone to overfitting when used directly for downstream task fine-tuning. After removing either of the two losses (G4/G5), the performance of the model also has a significant decrease. This shows that our design is effective and reasonable.
\begin{table}[]
\caption{The parameter size of {\model}. We report the ratio of trainable parameters and the time cost for 1 epoch pre-training.}
\label{table::params}
\begin{tabular}{ccccc}
\toprule
Dataset & \#Total Params & \#Trainable & Ratio & Time Cost \\
\midrule
Amazon & 418M & 0.14M & 0.33\% & 43.1s \\
Douban & 219M & 0.14M & 0.63\% & 24.6s \\
\bottomrule
\end{tabular}
\end{table}

\begin{figure}[]
  \centering
\includegraphics[width=0.9\linewidth]{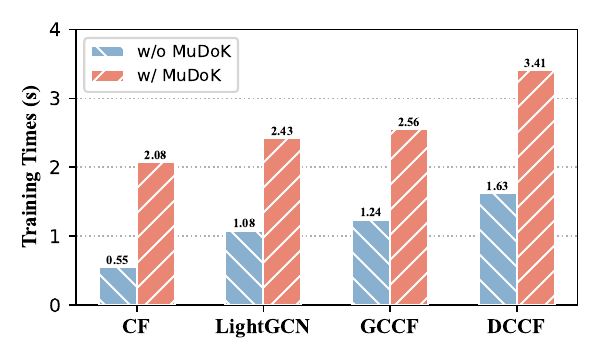}
  \vspace{-16pt}
  \caption{The efficiency analysis of {\model}. We report the training time on several recommendation backbones.}
  \label{figure::efficiency}
  \vspace{-16pt}
\end{figure}

\subsection{Efficiency Analysis (RQ4)}
To validate the efficiency of our method, we conduct efficiency analysis experiments. First, we quantify the number of parameters of our pre-trained model. As displayed in Table \ref{table::params}, we can find that the trainable parameters of {\model} are only a small percentage of the total parameters ($\leq$ 1\%). Training on large KGs of a million-scale is achievable within a minute using a single GPU, indicating the high efficiency of the {\pretrain} process in our model.

\par Meanwhile, we measure the latency that introduced our approach to downstream tasks. Figure \ref{figure::efficiency} reveals that {\prompt} does add some latency to the training of downstream tasks. However, this latency is tolerable given the overall enhancement in performance. The total training time of the model does not increase significantly.
\section{Conclusion}
In this paper, we focus on the problem of applying KGs in specific domains and diverse. In response to the problems of inefficient modeling, poor transferability, and inflexible application of downstream tasks in the existing research, we propose a method to realize collaborative pre-training of KGs as well as lightweight prefix prompt tuning. Besides, we construct a new KG pre-training and downstream application benchmark called {\dataset} for open-source and reproducible evaluation. The experimental results demonstrate the effectiveness across different domains and heterogeneous tasks. In the future, we think this work can be extended to a more unified framework domain-specific application and support more kinds of recommendation tasks like sequential recommendation and cross-domain recommendation.


\bibliographystyle{ACM-Reference-Format}
\bibliography{sample-base}


\end{document}